# Toward the Fully Physics-Informed Echo State Network - an ODE Approximator Based on Recurrent Artificial Neurons

Dong Keun Oh[a*]

[a] *Advanced Research Center, National Fusion Research Institute (NFRI), 169-148 Gwahak-ro, Daejeon 34133, South Korea*

ABSTRACT

Inspired by recent theoretical arguments, physics-informed echo state network (ESN) is discussed on the attempt to train a reservoir model absolutely in physics-informed manner. As the plainest work on such a purpose, an ODE (ordinary differential equation) approximator is designed to replicate the solution in sequence with respect to the recurrent evaluations. On the principal invariance of differential equations, the constraint in recurrence just takes shape to secure a proper regression method for the ESN-based ODE approximator. After then, the actual training process is established on the idea of two-pass strategy for regression. Aiming at the fully physics-informed reservoir model, a couple of nonlinear dynamical problems are demonstrated as the computations obtained from the proposed method in this study.

1. INTRODUCTION

   Echo state property is a fundamental concept to support the reservoir computing (RC) paradigm as a necessary condition for the "trainable readout mechanism" which enables the desired output from already configured recurrent neurons [1-3]. In such a neural computing model, the recurrent and input connections are initially created at random, and stay fixed, potentially being independent of the training process. To secure the desired output from a given input sequence, each evolution of the reservoir has to be at the state of "echoing" memory [1,4-6]. As a matter of logical inference, this property just means a passive state of the neurons accumulating the input history in a certain period toward the past [1]. In consequence, reservoirs in echo state are claimed to have such a memory fading away with evolution, as the past input is uniquely transformed into a high-dimensional space of the reservoir neurons [1,4].

Known for many insights to the neurobiological system with a plausible correspondence [7-8], for instance, to cerebral cognitive processes [9], the idea of echo state network (ESN) model has emerged with outperforming experiences in supervised learning for complicated dynamic patterns. To be practical, it has achieved great success in nonlinear system modeling, specifically, for generation [10], prediction [1,11-13], approximation [14], and adaptive controlling [15] of nonlinearly evolving sequences. At the same time, this approach has been recognized as a breakthrough to the difficulty in training of recurrent networks [1-3]. Hence, particularly related to the echo state property, reservoir models of nonlinear system have led specific interests in how the echo state memory, asymptotically fading out, surely assimilates the given chaotic behavior, or what conditions make such a target system structurally imposed to the trained readout to reproduce its underlying dynamics. Thus, in a number of different aspects, theoretical deductions have been calling out statistical (or stochastic) properties of the driving inputs to resolve such a notion to a matter of probability. For instance, a stricter definition of the echo state was introduced relying on non-autonomous input-driven dynamics [4], and a stronger inference is attained assuming the target system as an ergodic source [4,16-17]. On such a ground, the efficacy of ESN was ensured in terms of universal approximation to be valid under mild conditions [18]. Furthermore, as a complement of the argument, it was also discussed that an ESN surely embedded a structurally stable dynamical system into its reservoir [16-17]. As the recentest idea in those theoretical developments, imposing the dynamics was proven to be actualized as an $L2$ approximation, and to reproduce the future observation of the given dynamical system, by means of regularized least square regression [17].

While the recent arguments are quite evident how an ESN model works to secure its target system's (chaotic) trajectories, the actual inference about "learning dynamical behaviors" has an inherent limitation so far, because the main idea just relies on a sampled sequence, eventually, ruling out chances to involve the dynamics itself into the learning process. As a shortcoming on the attempt of theoretical demonstration, it was already indicated that the underlying dynamics of the target system has been ignored in ESN models [17]. Then, any alternative development is critically recommendable to integrate the input driver's dynamics into the reservoir training scheme. Hence, care has to be taken of the recently emerging ideas to consolidate physical information with the machine learning frameworks [19-26]. In particular, further investigations are to be led into the direction to involve the underlying dynamics, inspiring in the analogy of the aforementioned arguments [16-17].

In this article, an ESN-based ODE approximator is described as the first realization of a reservoir training scheme *absolutely* in physics-informed manner. Actually, solving an ODE in the form of $d\mathbf{y}/dt = \mathbf{f}(\mathbf{y})$ can be the plainest approach straightforward to a physics-informed ESN model of generative recurrence. In other words, a differential equation itself can be attributed as physical information to the training scheme approximating recurrent outputs to the solution in sequence.

In detail, there are certain steps to describe this study, particularly, on the implementation. At first, a minor revision out of the typical reservoir model is proposed in accordance with a common feature of the solution, where the output is represented at each period stepping ahead from the sequential input. On the other hand, regarding the intrinsic property of the error terms, nonlinear regression is invoked as an indispensable matter of this physics-informed ESN model. In addition to those principal considerations, a proper concept of physical error is established in respect of the recurrent structure of ESN. Actually, the idea is developed in the sense of piecewise connections between each step of the reservoir's output, following the fundamental idea on the causality and the invariance (Lie symmetry) of differential equations. As a result, an ESN-based solver of ODE is implemented in a two-pass strategy of neural solution relying on the iterative regressions with a novel formulation of physical error terms. They are discussed in order, and a few demonstrations are exemplified in step with a conclusion in this work.

## 2. BACKGROUND - NEURAL ODE APPROXIMATOR AS A PHYSICS-INFORMED ESN

Actually, there has been common effort on physics-driven machine learning, mostly, applied to the feedforward neural networks [19-23]. Such a concept has emerged on the aim of "learning analytic rules instead of data", and just derived a recent activity with the conceptual outline of "physics-informed neural network (PINN)" [20]. As shown in the studies such as a neural approximator for ordinary differential equations (ODEs) [22-23], their supervised learning process is typically driven by the deviation from the neural output's causal relations, namely, of the governing equations or its physical principles.

As an extension up to the recent efforts, a few pilot studies of "physics-informed ESN" are also found incorporated with physical error terms [24-26]. Nonetheless, in spite of such a pioneering effort, physical terms were employed just in part as a subsidiary contribution in addition to the conventional training scheme toward a sampled target sequence [24-25]. In contrast with the already reported examples of feedforward network, it is probably challenging to train a physics-driven ESN model without prerequisite data of the target system. Indeed, one has to point out the recurrent structure whose piecewise streaming is inevitably prone to be out of the target system's dynamical properties, and it is described later to reflect the invariance of differential equations in continuum. Under the circumstance, no meaningful development has not been found yet, not only to activate theoretical demonstrations of the embedding (or approximating) ability, but also to establish a pure physics-informed reservoir model as a foundation of those rigorous investigations, when the underlying dynamics is fully integrated into the reservoir.

Now, developing toward the fully physics-informed ESN, the basic idea of ESN is extended in a straightforward attempt to represent the ODE's solution, in which there is an aim of "machine learning by causality" after a simple differential equation $d\mathbf{y}/dt = \mathbf{f}(\mathbf{y})$.

## 3. OUTLINE - ESN-BASED ODE APPROXIMATOR

An ESN model of ODE solution is conceived in a presentative way according to the sequential nature of recurrent evaluation. In order to represent such an approximator, the reservoir's readout has to follow discrete steps of the given ODE with respect to the independent variable $t$ getting forward at a certain interval, i.e., to $t + \Delta t$ in sequence. Thus, by means of a few modifications from the standard layout, it is possible to take an appropriate idea to update the $(n+1)$th recurrent state $\mathbf{h}^{n+1}$ of reservoir neurons. In the following manner (Eq. 1), the evolutions are represented with respect to $(\mathbf{y}^n, \tau^n)$, i.e., as the $n$th input vector in case of $d\mathbf{y}/dt = \mathbf{f}(\mathbf{y})$, or with respect to $(\mathbf{y}^n, t^n, \tau^n)$ in case of $d\mathbf{y}/dt = \mathbf{f}(\mathbf{y}, t)$,

$$\mathbf{h}^{n+1} = \sigma(\boldsymbol{\omega} \cdot \mathbf{h}^n + \mathbf{v} \cdot \mathbf{y}_0^n + \mathbf{b}\ \tau^n + \mathbf{c})$$
$$\text{or } \mathbf{h}^{n+1} = \sigma(\boldsymbol{\omega} \cdot \mathbf{h}^n + \mathbf{v} \cdot \mathbf{y}_0^n + \mathbf{a}\ t^n + \mathbf{b}\ \tau^n + \mathbf{c}) \tag{1}$$

where $\sigma$ means element-wise operation of the nonlinear activation, and the variables **ω, v, a, b** and **c** are the weight matrices of the hidden layer's (the reservoir's) connection randomly configured in the beginning [1-3].

Then, the readout $\mathbf{y}^{n+1}$ comes out of the update of hidden state as the following,

$$\mathbf{y}^{n+1} = \mathbf{y}_0^n + \tau^n \, \mathbf{w} \cdot \mathbf{h}^{n+1} \tag{2}$$

where **w** is the readout matrix of weights to be determined by means of regression, i.e., in the training process.

Indeed, Eq. 2 describes an incremental update, just aiming at being to conform to the integral of an ODE $d\mathbf{y}/dt = \mathbf{f}(\mathbf{y})$ or $d\mathbf{y}/dt = \mathbf{f}(\mathbf{y},t)$ from $t = t^n$ along the interval from $t^n$ to $t^n + \tau^n$. Thus, it is suitable to express the solution after $\tau^n$, namely at $t^n + \tau^n$, where the initial condition is given by the $n^{\text{th}}$ input $\mathbf{y}_0^n$. In practice, the intervals are assigned to be constant, for simplicity on the attempts of this study for demonstration.

Following up of the logical layout in Eq. 1 and 2, one has to go into the details of regression to the given ODE's solution. Actually, the matrix **w** will be determined for an approximation to the solution sampled by the intervals. To be sure, the *t*-derivative term $d\mathbf{y}/dt$ must be informed in this process of training, and the proposed layout (Eq. 1 and 2) is possible to provide the explicit differential from the hidden state thanks to the dependence on the interval $\tau^n$. Hence, the ESN-based ODE approximator is thought on the ground of such an outlining.

In spite of the minor updates, there is an essential discrimination in the actual training steps to be compared with the typical method of regularized linear regression. In practice, linear regression can be applied [1-3], in terms of squared errors of $\left|\mathbf{y}_{\text{target}}^n - \mathbf{y}^n(\mathbf{w})\right|^2$, to the fixed target of sampled data $\mathbf{y}_{\text{target}}^n$, since $\mathbf{y}_{\text{target}}^n - \mathbf{y}^n(\mathbf{w})$ is nothing but a linear relation ruled by **w** owing to the explicitly prepared hidden states; it is illustrated in Fig. 1. However, an intrinsic difficulty will be brought as a matter of course, whenever nonlinearity on the right-hand-side of $\mathbf{f}(\mathbf{y})$ or $\mathbf{f}(\mathbf{y},t)$ is involved into the least square regression. Just at a glance to a particular formula $\left|d\mathbf{y}/dt - \mathbf{f}\right|^2$ after the causality of the ODE, it is simple to notice that, even in such a straightforward idea of physical information

[20,22-25], nonlinearity out of $\mathbf{f}(\mathbf{y})$ or $\mathbf{f}(\mathbf{y},t)$ has to invoke an essential measures of iterative regression which is not avoidable in case of the nonlinear least square minimization. Thus, this matter is a critical hurdle to be taken into account in the design of training scheme.

On the other hand, as implied in the formulation, any attempt on the outline of this study has to respect the conceptual sense of sampling, as illustrated in Fig. 2, on the real solution. On this ground, the $\mathbf{y}^n$ just represents a numerical approximation to the solution for each period. Ultimately, the reservoir model works in autonomous mode to generate the solution in sequence. Thus, all readouts are to be connected in series with each other as an approximation to the solution. In other words, at each initial condition given by $\mathbf{y}_0^n$ of the $n^{\text{th}}$ input, the incremental part of the readout $\Delta^n$ is to be consistent with the solution of $d\mathbf{y}/dt = \mathbf{f}(\mathbf{y})$ with respect to each initial condition $\mathbf{y}_0^n$.

$$\begin{cases} \Delta^{n+1}(\tau^n, \mathbf{y}_0^n; \mathbf{w}) = \tau^n \, \mathbf{w} \cdot \mathbf{h}^{n+1} \\ \mathbf{h}^{n+1} = \sigma(\boldsymbol{\omega} \cdot \mathbf{h}^n + \mathbf{v} \cdot \mathbf{y}_0^n + \mathbf{b} \, \tau^n + \mathbf{c}) \end{cases} \tag{3}$$

In consequence, the readout can be described as a series of previous $\Delta^n$ according to the recurrent evaluation, when the $n^{\text{th}}$ readout $\mathbf{y}^n$ is transferred to the next step as an input $\mathbf{y}_0^n$, i.e., in autonomous manner.

$$\begin{aligned} \mathbf{y}^{n+1} &= \mathbf{y}^n + \Delta^{n+1}(\tau^n, \mathbf{y}^n; \mathbf{w}) \\ \Rightarrow \mathbf{y}^{n+1} &= \mathbf{y}^{n-n_0} + \Delta^{n+1} + \Delta^n + \ldots + \Delta^{n-n_0+1} \quad (n \geq n_0 \geq 0) \end{aligned} \tag{4}$$

Meanwhile, the fundamental logic of differential equation is always satisfied in the flashback (Eq. 4) to the point $n_0$ shifted backward. Namely, the output $\mathbf{y}^n$ is also the same solution regardless of what $n_0$ is selected for the initial state. Hence, one must take into account such a nature of the solution as an important consideration to establish the details. Actually, it is noted as the most particular idea in this study of physics-informed ESN,to take care of a consistency in the sequential readouts regarding the general principle of differential equations; it is described in the next section.

4. SOLUTIONS IN RECURRENCE - HOW TO CONSTRAIN THE NEURAL OUTPUT

The Lie invariance is the neatest approach to discuss the successive evaluations as a solution in the form of recurrent sequence, which is the fundamental idea about general property of differential equations, particularly, in terms of arbitrary shifts by a certain parameter [27-28]. Going a step further into the details of training, one has to consider a loss function to minimize, at first, taking into account the causality at each prompt response on the interval. At the same time, a distinctive idea on the recurrent evaluation just inspires another important constraining by means of the strict relation *between the readouts* to each other.

To describe such a particular concept on the differential equations, every readout of the reservoir can be denoted by $\mathbf{y}^*$ as a solution of $d\mathbf{y}^*/dt^* = \mathbf{f}(\mathbf{y}^*, t^*)$ on the evolution toward $t^* = t + \tau$ from the moment $t$ with the initial value $\mathbf{y}$. Then, the Lie transformation is naturally introduced to correspond $(\mathbf{y}, t)$ to $(\mathbf{y}^*, t^*)$, i.e., considering the both of independent and dependent variables.

$$\mathbf{y}^* = \mathbf{Y}(\mathbf{y}, t; \tau), \quad t^* = T(\mathbf{y}, t; \tau) = t + \tau$$
$$\Xi = \mathbf{f}(\mathbf{y}, t) \cdot \frac{\partial}{\partial \mathbf{y}} + \frac{\partial}{\partial t} \tag{5}$$

As written in Eq. 5, the transformation is ruled by the parameter $\tau$, and this logical layout is supported by the infinitesimal evolution operator $\Xi$. Namely, as a special case of the general transformation, $T(\mathbf{y}, t; \tau) = t + \tau$ is able to be introduced, and $\mathbf{Y}(\mathbf{y}, t; \tau)$ is represented as $\mathbf{y} + \tau (\Xi \cdot \mathbf{y}) + O(\tau^2) \approx \mathbf{y} + \tau \mathbf{f}(\mathbf{y}, t)$ in terms of infinitesimal evolution of $\tau$ [27-28]. Thus, the constraint equation (Eq. 6) is derived by means of the general differential formula in the Lie's theory of differential equation, where Eq. 5 is proposed to make it conform to the evolution getting to $\mathbf{y}^*$ over the period of $\tau$ [27-28].

$$\frac{d\mathbf{y}^*}{dt^*} = \frac{D_t \mathbf{Y}}{D_t T} = \frac{(\frac{\partial}{\partial t} + \dot{\mathbf{y}} \cdot \frac{\partial}{\partial \mathbf{y}}) \mathbf{Y}}{(\frac{\partial}{\partial t} + \dot{\mathbf{y}} \cdot \frac{\partial}{\partial \mathbf{y}}) T}$$
$$\Rightarrow \mathbf{f}(\mathbf{y}^*, t^*) = \left( \frac{\partial}{\partial t} + \mathbf{f}(\mathbf{y}, t) \cdot \frac{\partial}{\partial \mathbf{y}} \right) \mathbf{y}^* \tag{6}$$

In accordance with the idea, the formulation in Eq. 7 conceptually fits the sequential output. Indeed, it is introduced as a solution virtually evolving along the period of $\tau^n$, where $(\mathbf{y}^n, t^n)$ is given to be the initial state.

$$\begin{cases} \mathbf{y}^{n+1} = \mathbf{y}^n + \tau^n \Delta^{n+1}(\mathbf{y}^n, t^n; \tau^n) \\ t^{n+1} = t^n + \tau^n \end{cases} \quad (7)$$

To be consistent with the equation $d\mathbf{y}/dt = \mathbf{f}(\mathbf{y}, t)$, the essential constraint on $\mathbf{y}^{n+1}$, which is ruled by $\Delta^{n+1}$, should be applied above all in terms of the $\tau$-derivative (Eq. 8-1 and 8-2). In addition, another primary constraint should be taken following the invariance principle represented in Eq. 6, for the recurrence between the output and the input (or the previous output) of the reservoir (Eq. 8-3).

$$\Delta^{n+1}(\mathbf{y}^n, t^n, \tau^n) + \tau^n \partial \Delta^{n+1}(\mathbf{y}^n, t^n, \tau^n)/\partial \tau = \mathbf{f}(\mathbf{y}^{n+1}, t^{n+1}) \quad (\text{for } \tau = \tau^n) \quad (8\text{-}1)$$

$$\Delta^{n+1}(\mathbf{y}^n, t^n, 0) = \mathbf{f}(\mathbf{y}^n, t^n) \quad (\text{for } \tau = 0) \quad (8\text{-}2)$$

$$\left( \mathbf{f}(\mathbf{y}^n, t^n) \cdot \frac{\partial}{\partial \mathbf{y}^n} + \frac{\partial}{\partial t^n} \right) \mathbf{y}^{n+1} = \mathbf{f}(\mathbf{y}^{n+1}, t^{n+1}) \quad (8\text{-}3)$$

As a result, the constraint equations in Eq. 8 are obtained to make the readout tight enough as an approximation to the sequential solution getting into step with its recurrent evaluation. On the basis of Eq. 8, the error terms will be specified for a loss function of regression to be applied to the actual process of training.

It is worthy to note that there has been a strong inspiration regarding the Lie invariance to generate the numerical solution in a recurrent manner [29-30], in spite of little relevance to the idea of general neural architectures to replicate the behavior of arbitrary targets. Actually, the previous studies of such a recurrent structure just relied on the Lie transformation as they are implemented by the expansion of Kronecker powers [29-30]. Even being termed a neural network, such a truncated series is an explicit expression in the operator formulation itself from $\mathbf{f}(\mathbf{y}, t)$ [30-31]. Hence, the $(\mathbf{y}, t)$- constraint between each contiguous step (Eq. 8-3) is understood in line of the Lie invariance on the recurrent formulation, which just gives an insight in confidence to bind the reservoir's readouts into a causally related sequence according to the ODE.

## 5. TWO-PASS STRATEGY - THE ACTUAL SCHEME OF REGRESSIONS

In the commonest workflow of ESN, the benefit of fixed input sequence is paramount, since a simple regression just makes enough to train the reservoir model owing to the virtually separated output channel from complicated online feedback in recurrence [1-3]. Furthermore, such a frozen state, prepared in the phase of training (Fig. 1), just ensures the final usage of recurrent exploitation; actually, it is tricky as the same sequence is assigned to the input one step earlier than the target, making it consistent with the autonomous generation later on. Thus, it is recommendable to keep such an advantage as much as possible for this study also.

However, in case of ODE approximator based on the recurrent layout in Eq. 1 ~ 4, the input sequence is unknown, in principle, to be determined as an accurate solution. Thus, it does not makes sense either to prepare a proper input at one go, or to keep up with the approximation without spoiling the straightforwardness of regression. Out of such an obstacle, the idea of training scheme just emerges working in two stages by means of a kind of precasting with a trial solution. Once an accurate approximation is secured based on the trial solution at the first stage, it is possible to prepare the input sequence for the next stage on the ultimate purpose to generate the solution in recurrent manner. As illustrated in Fig. 3, one can understand the idea schematically to actualize the two-pass regression scheme in terms of least square minimization ruled by the differential equation.

Once the solution strategy is established, each stage of regression is implemented to minimize the sum of square errors which indicate how much the approximation deviates regarding the causality described by the constraint equations. To build such a function of square errors, the error vectors are formulated with respect to Eq. 8 not only for each step (Eq. 8-1 and 8-2), but also between each contiguous evaluation (Eq. 8-3). Actually, one is able to regard the constraint equations in Eq. 8 as a general formulation, and the *t*-derivative term in Eq. 8-3 is neglected just because of the equations in the form of $d\mathbf{y}/dt = \mathbf{f}(\mathbf{y})$; however, the formulation is easy to extend to the cases of absolute *t*-dependence in $\mathbf{f}$.

At the first stage, the trial solution is denoted by $\mathbf{y}_0^n$. Then, to obtain the neural approximation in sequence, $\overline{\mathbf{y}}^{n+1}$ is supposed to be the reservoir's readout in the next step evolving from $\overline{\mathbf{y}}^n$ at present.

Thus, one has to take into account the previous readout as the initial value of the next step, whereas the second stage, even at the same logic, must bring the input sequence instead.

$$\bar{\mathbf{y}}^{n+1} = \bar{\mathbf{y}}^n + \tau^n \bar{\mathbf{w}} \cdot \bar{\mathbf{h}}^{n+1},$$
$$\text{where } \begin{cases} \bar{\mathbf{h}}^{n+1} = \sigma(\bar{\mathbf{z}}^{n+1}) \\ \bar{\mathbf{z}}^{n+1} = \mathbf{b}\tau^n + \boldsymbol{\omega} \cdot \bar{\mathbf{h}}^n + \mathbf{v} \cdot \mathbf{y}_0^n + \mathbf{c} \end{cases} \quad (9)$$

Thus, the second stage output $\mathbf{y}^n$ is represented in terms of $\bar{\mathbf{y}}^n$ out of the first stage.

$$\mathbf{y}^{n+1} = \bar{\mathbf{y}}^n + \tau^n \mathbf{w} \cdot \mathbf{h}^{n+1},$$
$$\text{where } \begin{cases} \mathbf{h}^{n+1} = \sigma(\mathbf{z}^{n+1}) \\ \mathbf{z}^{n+1} = \mathbf{b}\tau^n + \boldsymbol{\omega} \cdot \mathbf{h}^n + \mathbf{v} \cdot \bar{\mathbf{y}}^n + \mathbf{c} \end{cases} \quad (10)$$

Hence, the error vectors of the first stage is defined as the following, according to the three constraint equations in Eq. 8,

$$\bar{\mathbf{e}}_1^{n+1} = \mathbf{f}(\bar{\mathbf{y}}^{n+1}) - \bar{\mathbf{w}} \cdot \left\{ \sigma(\bar{\mathbf{z}}^{n+1}) + \tau^n \mathbf{b} \dot{\sigma}(\bar{\mathbf{z}}^{n+1}) \right\}$$
$$\bar{\mathbf{e}}_2^{n+1} = \mathbf{f}(\bar{\mathbf{y}}^n) - \bar{\mathbf{w}} \cdot \sigma(\bar{\mathbf{z}}_0^{n+1}) \quad (11\text{-}1)$$
$$\bar{\mathbf{e}}_3^{n+1} = \left( \mathbf{f}(\bar{\mathbf{y}}^n) \cdot \frac{\partial}{\partial \bar{\mathbf{y}}^n} \right) \bar{\mathbf{y}}^{n+1} - \mathbf{f}(\bar{\mathbf{y}}^{n+1})$$

$$\mathbf{e}_1^{n+1} = \mathbf{f}(\mathbf{y}^{n+1}) - \mathbf{w} \cdot \left\{ \sigma(\mathbf{z}^{n+1}) + \tau^n \mathbf{b} \dot{\sigma}(\mathbf{z}^{n+1}) \right\}$$
$$\mathbf{e}_2^{n+1} = \mathbf{f}(\mathbf{y}^n) - \mathbf{w} \cdot \sigma(\mathbf{z}_0^{n+1}) \quad (11\text{-}2)$$
$$\mathbf{e}_3^{n+1} = \left( \mathbf{f}(\mathbf{y}^n) \cdot \frac{\partial}{\partial \bar{\mathbf{y}}^n} \right) \mathbf{y}^{n+1} - \mathbf{f}(\mathbf{y}^{n+1})$$

and the second stage also goes in the similar manner, where $\bar{\mathbf{z}}_0^{n+1}$ is $\boldsymbol{\omega} \cdot \bar{\mathbf{h}}^n + \mathbf{v} \cdot \mathbf{y}_0^n + \mathbf{c}$, and $\mathbf{z}_0^{n+1}$ is $\boldsymbol{\omega} \cdot \mathbf{h}^n + \mathbf{v} \cdot \bar{\mathbf{y}}^n + \mathbf{c}$ meaning $\tau = 0$ on the constraint of Eq. 8-2. Being notable to the first stage, the third constraint (Eq. 8-3) requires the derivative by $\bar{\mathbf{y}}^n$ on which the hidden state $\bar{\mathbf{h}}^{n+1}$ doesn't have any explicit dependence. Thus, care has to be taken of the $\bar{\mathbf{y}}^n$- derivative term $\partial \bar{\mathbf{y}}^{n+1}/\partial \bar{\mathbf{y}}^n$ to be imbedded into the error vector $\bar{\mathbf{e}}_3$ in an implicit manner by means of the identities of $\partial \bar{\mathbf{y}}^{n+1}/\partial \bar{\mathbf{h}}^n = \left(\partial \bar{\mathbf{y}}^{n+1}/\partial \bar{\mathbf{y}}^n\right) \cdot \left(\partial \bar{\mathbf{y}}^n/\partial \bar{\mathbf{h}}^n\right)$ and $\partial \bar{\mathbf{y}}^n/\partial \bar{\mathbf{h}}^n = \mathbf{w}$.

After all, the error vectors are written with tensorial indices just for code-ready condition, introducing the loss functions of square error, i.e., $\bar{L} = \bar{\mathbf{e}}_1^T \cdot \bar{\mathbf{e}}_1 + \bar{\mathbf{e}}_2^T \cdot \bar{\mathbf{e}}_2 + \bar{\mathbf{e}}_3^T \cdot \bar{\mathbf{e}}_3$ of the first stage and $L = \mathbf{e}_1^T \cdot \mathbf{e}_1 + \mathbf{e}_2^T \cdot \mathbf{e}_2 + \mathbf{e}_3^T \cdot \mathbf{e}_3$ of the second stage,

$$[\bar{\mathbf{e}}_1^{n+1}]_i = f_i([y_{0j}^0 + \sum_m \bar{w}_{jm}(\tau^1 \sigma_m^1 + \tau^2 \sigma_m^2 + ... + \tau^n \sigma_m^{n+1})]) - \sum_m \bar{w}_{im}\left(\sigma_m^{n+1} + \tau^n b_m \dot{\sigma}_m^{n+1}\right)$$
$$[\bar{\mathbf{e}}_2^{n+1}]_i = f_i([y_{0j}^0 + \sum_m \bar{w}_{jm}(\tau^1 \sigma_m^1 + \tau^2 \sigma_m^2 + ... + \tau^{n-1} \sigma_m^n)]) - \sum_m \bar{w}_{im} \sigma_{0m}^{n+1} \qquad (12\text{-}1)$$
$$[\bar{\mathbf{e}}_3^{n+1}]_i = \sum_{l,m} \frac{\tau^n}{\tau^{n-1}} \bar{w}_{im} \omega_{ml} \dot{\sigma}_l^{n+1} \sigma_{0m}^{n+1} - \sum_m \bar{w}_{im}\left(\sigma_m^{n+1} + \tau^n b_m \dot{\sigma}_m^{n+1} - \sigma_{0m}^{n+1}\right)$$

$$[\mathbf{e}_1^{n+1}]_i = f_i([\bar{y}_j^n + \tau^n \sum_m w_{jm} \sigma_m^{n+1}]) - \sum_m w_{im}\left(\sigma_m^{n+1} + \tau^n b_m \dot{\sigma}_m^{n+1}\right)$$
$$[\mathbf{e}_2^{n+1}]_i = f_i([\bar{y}_j^n]) - \sum_m w_{im} \sigma_{0m}^{n+1} \qquad (12\text{-}2)$$
$$[\mathbf{e}_3^{n+1}]_i = \sum_{j,l,m} \tau^n w_{im} v_{jm} w_{jl} \dot{\sigma}_l^{n+1} \sigma_{0m}^{n+1} - \sum_m w_{im}\left(\sigma_m^{n+1} + \tau^n b_m \dot{\sigma}_m^{n+1} - \sigma_{0m}^{n+1}\right)$$

where $\sigma_m^{n+1} = [\sigma(\mathbf{z}^{n+1})]_m$ and $\sigma_{0m}^{n+1} = [\sigma(\mathbf{z}_0^{n+1})]_m$.

As a result, each stage of least square regression is ready in action relying on the loss function, to determine the weight matrices $\bar{\mathbf{w}}$ and $\mathbf{w}$ for each stage in turn. Solving regularized linear equations based the Gauss-Newton method, the local minimum of each loss function is obtained from the iterative process updating the weight matrix $\bar{\mathbf{w}}$ or $\mathbf{w}$ in the following manner [32],

$$\mathbf{w} \to \mathbf{w} + ([\mathbf{J}]^T \cdot [\mathbf{J}] + \lambda \mathbf{I})^{-1} \cdot [\mathbf{J}]^T \cdot \mathbf{e}$$
$$, \text{where } \begin{cases} [\mathbf{J}]_{i,jk} = \partial e_i / \partial w_{jk} \\ \mathbf{e} = \begin{bmatrix} \mathbf{e}_1 \\ \mathbf{e}_2 \\ \mathbf{e}_3 \end{bmatrix} \end{cases} \qquad (13)$$

where $\lambda$ is the Tikhonov regularization [33] parameter adjusted to be small enough.

It is remarkable that, even though iterative regression is not avoidable mainly because of the nonlinearity potentially involved in $\mathbf{f}(\mathbf{y})$, the simple outline of least square minimization is still valid, principally identical to the typical reservoir model, owing to the passive status prepared by the input sequence. Furthermore, except for such a potential nonlinearity of $\mathbf{f}(\mathbf{y})$, there are quadratic terms of $\mathbf{w}$ in $\mathbf{e}_3$, which is the only intrinsic nonlinearity. In this condition, one is able to think about separating the linear terms to make it more effective than this state of preliminary attempt [34]. On the other hand, there seems to be another chance to improve the regression scheme, for instance, employing the Levenberg-Marquardt algorithm [35].

## 6. IMPLEMENTATION - TEST OF HARMONIC OSCILLATOR

As the details of regression are arranged, it is tested above all whether the implemented approximator works as planned. Namely, a pilot program is carried out to get a solution of the simple harmonic oscillator, i.e., $\dot{y}_1 = y_2$ and $\dot{y}_2 = -y_1$. The reservoir neurons are just built based on the custom class, importing the Tensorflow library (1.13.1), with respect to the proposed structure. Evaluated in the Tensorflow session, the hidden states evolving in recurrence are prepared as a Numpy array, and applied to the regression process, which is implemented using the simple linear solver, written in plain Python.

Setting up the ESN model, 200 neurons of the hidden (recurrent) layer are configured at random connection whose connectivity is 0.01~0.1, and the spectral width of the hidden connection's weight $\boldsymbol{\omega}$ is controlled by its 2-norm to be 10.0, as the 2-norm is the upper bound of the spectral radius of $\boldsymbol{\omega}$. Notably, the large spectral radius doesn't belong to the sufficient condition for echo state in the conventional setup ($\rho(\boldsymbol{\omega}) < 1$) [1], but just selected to be good enough by trial-and-errors in this special modification. Thus, it is interesting to investigate the echo state property of this system in accordance with its specific structure similar to the residual network, which can be carried out in the future.

For the main part of regression, a trial solution is prepared, by means of the Euler method, on the sequential steps of $\tau = 0.05$, and is cropped to get 500 points on the span of neural solution. In this course, 150 points of the leading part are dropped to wash out the initial transient of reservoir [3]. Then, the regression process of the first stage is invoked to carry out the Gauss-Newton iteration which needs the initial guess of $\bar{\mathbf{w}}^{(initial)}$. Finding nothing for alternative, the initial guess $\bar{\mathbf{w}}^{(initial)}$ is selected just to replicate the trial solution, i.e., to make $\mathbf{y}_0^n + \tau^n \bar{\mathbf{w}} \cdot \bar{\mathbf{h}}^{n+1}$ equal to $\mathbf{y}_0^{n+1}$. As obtained in a different context from the actual formulation (Eq. 9), the first stage output with $\bar{\mathbf{w}}^{(initial)}$ can be far from the expectation as demonstrated in Fig. 4. After the iterations, the final $\bar{\mathbf{y}}^n$ comes out to be accurate enough (Fig. 4). Then, the result is transferred to the second stage of regression. Because the harmonic oscillator is a linear equation, it is quite easy to converge so that only a few iterations are enough to stabilize the relative change of total loss $|\Delta L/L|$ to be less than $10^{-5}$ in the both stages.

As a result, 10 iterations are applied for the both stages of the two-pass regression. Thus, the result is obtained with the maximum error deviated from the true solution as about $5 \times 10^{-3}$, as presented in Fig. 5 and 6; the figures display the differences from the true solution for each stage as well as the square norm of error vectors after the successive stages of regression. Eventually, the demonstration is completed to verify the regression process which is proposed as a training scheme of the ESN-based ODE approximator.

## 7. RESULT - A FEW DEMONSTRATIONS OF NONLINEAR DYNAMICS

After the verification is successful, the cases of nonlinear differential equation are attempted as presented in this section; one is the van der Pol oscillator [36], and the other is the Lorentz chaotic system [37].

### 7.1 Case 1; the van der Pol oscillator

The van der Pol equation $\ddot{y} + (y^2 - 1)\dot{y} + y = 0$ can be written in the form of $d\mathbf{y}/dt = \mathbf{f}(\mathbf{y})$, where $\mathbf{f}(\mathbf{y}) = (y_2, y_2 - y_1 - y_1^2 y_2)$ for $\mathbf{y} = (y_1, y_2)$ [30,36]. The reservoir of 300 neurons are configured in the same manner of the previous verification. For the evolution span of 500 points, the input sequence is prepare by the trial solution with $\tau = 0.1$. In the both stages of regression, the regularization parameter $\lambda = 10^{-7}$ is assigned, and the iterations are carried out as presented in Fig. 7 and Fig. 8. As a result, the neural solution is obtained comparing with the trial solution for the first stage (Fig. 9).

## 7.2 Case 2; the Lorenz system

The Lorenz equations are well-known as a nonlinear system of chaotic behavior. The system of differential equations, which are $\dot{y}_1 = \sigma(y_2 - y_1)$, $\dot{y}_2 = y_1(\rho - y_3) - y_2$ and $\dot{y}_3 = y_1 y_2 - \beta y_3$, exhibit chaotic attractors, which Lorenz discovered with the parameters $\sigma = 10$, $\rho = 28$, and $\beta = 8/3$ [37]. To generate the 200 sequential solutions on the intervals of $\tau = 0.03$, the ESN-based ODE approximator is configured with 1500 hidden neurons which is large amount selected in reason after many struggles to make the regression converged. To prepare the trial solution, the time interval $\tau/20$ is applied to the Euler method, and the original period $\tau$ is recovered by means of downsampling; it is required to make the Euler step fine enough, since this model is checked to be vulnerable against the error of the trial solution more than the previous examples. After then, the regression is done for each stage, and its progress in detail is described in Fig. 8 and Fig. 9 respectively per stage. Eventually, the neural solution is obtained, as shown in Fig. 10.

## 8. DISCUSSION

As it has been recognized, the feedforward models of physics informed neural network are just supposed to learn the solution at a specified condition of initial or boundary constraints, which means that one has to do training repeatedly with respect to the changing conditions. Thus, it is not suitable to implement a neural model to replicate the target system's behavior under the various situations in practice. In such a point of view, the ESN-based approximator is important to cover such a shortcoming owing to the generative concept based on the reservoir architecture which is expected to be capable of assimilating chaotic nonlinear trajectories [16-18]. On the other hand, its applicability is still questionable because of the performance at the required computational cost of iterations, according to the examinations in this work.

Nonetheless, its power of insight is possibly extended to the theoretical discussions on the dynamical behavior hidden in the reservoir-based model of nonlinear trajectories. Paying attention to the logical layout to be consistent with the theory of differential equations, this study just implements the neural ODE approximator, in particular, evaluating in step respecting the integration along each interval of the dynamical interactions. What is the most remarkable is that the physical information is imposed between each recurrence of the reservoir by means of the special constraint of invariant principle. As the differential equation itself doesn't prompt the explicit dependence on the initial conditions, the effect between each step is considered based on the inventive approach to be essential in the ESN-based ODE model. Thus, the dynamical property embedded into the reservoir's recurrences can be investigated in the context of invariance principles from the given physical information, and this approach may give a critical hint as a refreshing point to the concepts of physics-informed ESN. For a practical activity, the idea of this study is also able to enhance the previous attempts of the physics-informed ESN, for instance, recovering hidden states' behavior by means of this style of constraint between the recurrent evolution [25].

## 8. CONCLUSION

An ODE approximator of ESN-based architecture is successfully demonstrated on the attempt to implement the first fully physics-informed training scheme for the reservoir computing model. The invariance of differential equations just establishes an essential idea to constrain the recurrent models in terms of the given ODE. In consequence, some nonlinear trajectories are solved by means of the (recurrent) neural approximation, as a demonstration to verify how the implemented courses work.

## ACKNOWLEDGEMENT

This work was supported by the Korean Ministry of Science, ICT.

## REFERENCES

1. Jaeger, H. (2001). "The "echo state" approach to analysing and training recurrent neural networks." Technical Report GMD Report 148, German National Research Center forInformation Technology.
2. Lukoševičius M., Jaeger H. "Reservoir computing approaches to recurrent neural network training." Computer Science Review, 3 (2009), pp. 127-149
3. Lukoševičius M. "A practical guide to applying echo state networks." Neural networks: Tricks of the trade, Springer (2012), pp. 659-686

4. Manjunath, G., and Jaeger, H. "Echo state property linked to an input: Exploring a fundamental characteristic of recurrent neural networks." Neural computation, 25(3) (2013), pp. 671-696

5. Buehner, M., and Young, P. "A tighter bound for the echo state property." IEEE Transactions on Neural Networks, 17(3) (2006), pp. 820-824.

6. Yildiz, I.B., Jaeger, H., and Kiebel, S.J. "Re-visiting the echo state property." Neural Networks, 35 (2012), pp. 1-9

7. Tanaka, G., Yamane, T., Héroux, J. B., Nakane, R., Kanazawa, N., Takeda, S., ... and Hirose, A. "Recent advances in physical reservoir computing: A review." Neural Networks, 115 (2019), pp. 100-123.

8. Gürel, T., Rotter, S., and Egert, U. "Functional identification of biological neural networks using reservoir adaptation for point processes." Journal of computational neuroscience, 29(1-2) (2010), pp. 279-299.

9. Dominey P.F. "Complex sensory-motor sequence learning based on recurrent state representation and reinforcement learning Biological Cybernetics." 73 (1995), pp. 265-274; Dominey P., Arbib M., Joseph J.-P. "A model of corticostriatal plasticity for learning oculomotor associations and sequences." Journal of Cognitive Neuroscience, 7 (1995), pp. 311-336

10. Hauser H., Ijspeert A.J., Füchslin R., and Maass W. "The role of feedback in morphological computation with compliant bodies." Biological Cybernetics, 106 (2012), pp. 595-613

11. Jaeger H. "Adaptive nonlinear system identification with echo state networks." Advances in neural information processing systems (2003), pp. 609-616

12. Pathak, J., Hunt, B., Girvan, M., Lu, Z., and Ott, E. "Model-free prediction of large spatiotemporally chaotic systems from data: A reservoir computing approach." Physical review letters, 120(2) (2018), 024102.

13. Lu, Z., Pathak, J., Hunt, B., Girvan, M., Brockett, R., and Ott, E. "Reservoir observers: Model-free inference of unmeasured variables in chaotic systems." Chaos: An Interdisciplinary Journal of Nonlinear Science, 27(4) (2017), 041102.

14. Bertschinger N., and Natschläger T. "Real-time computation at the edge of chaos in recurrent neural networks Neural Computation." 16 (2004), pp. 1413-1436

15. Jaeger, Herbert, and Harald Haas. "Harnessing nonlinearity: Predicting chaotic systems and saving energy in wireless communication." Science 304.5667 (2004) pp. 78-80.

16. Hart, A., Hook, J., and Dawes, J. "Embedding and approximation theorems for echo state networks." Neural Networks 128 (2020), pp. 234-247 .

17. Hart, A. G., Hook, J. L., and Dawes, J. H. "Echo State Networks trained by Tikhonov least squares are $L^2$ approximators of ergodic dynamical systems." arXiv preprint arXiv: 2005.06967 (2020).

18. Grigoryeva, L., and Ortega, J. P. "Echo state networks are universal." Neural Networks, 108 (2018), pp. 495-508.

19. King, R., Hennigh, O., Mohan, A., and Chertkov, M. "From deep to physics-informed learning of turbulence: Diagnostics." arXiv preprint arXiv:1810.07785 (2018).

20. Raissi, M., Perdikaris, P., and Karniadakis, G. E."Physics-informed neural networks: A deep learning framework for solving forward and inverse problems involving nonlinear partial differential equations." Journal of Computational Physics, 378 (2019), pp. 686-707.

21. Tartakovsky, A. M., Marrero, C. O., Perdikaris, P., Tartakovsky, G. D., and Barajas-Solano, D. "Physics-Informed Deep Neural Networks for Learning Parameters and Constitutive Relationships in Subsurface Flow Problems." Water Resources Research, 56(5) (2020), e2019WR026731.

22. Filici, C. "On a neural approximator to ODEs." IEEE transactions on neural networks, 19(3) (2008), pp. 539-543.

23. Lagaris, I. E., Likas, A., and Fotiadis, D. I. "Artificial neural networks for solving ordinary and partial differential equations." IEEE transactions on neural networks, 9(5) (1998), pp. 987-1000.

24. Doan, N. A. K., Polifke, W., and Magri, L. "Physics-informed echo state networks for chaotic systems forecasting." In International Conference on Computational Science (2019, June) pp. 192-198. Springer, Cham.

25. Doan, N. A. K., Polifke, W., and Magri, L. "Learning hidden states in a chaotic system: A physics-informed echo state network approach." arXiv preprint arXiv:2001.02982 (2020).

26. Huhn, F., and Magri, L. "Learning ergodic averages in chaotic systems." arXiv preprint arXiv: 2001.04027 (2020).

27. Oliveri, F. "Lie symmetries of differential equations: classical results and recent contributions." Symmetry, 2(2) (2010), pp. 658-706.

28. Ibragimov, N. K. "Group analysis of ordinary differential equations and the invariance principle in mathematical physics (for the 150th anniversary of Sophus Lie)." Russian Mathematical Surveys, 47(4) (1992), pp. 89.

29. Ivanov, A., Iben, U., and Golovkina, A. "Physics-based polynomial neural networks for one-short learning of dynamical systems from one or a few samples." arXiv preprint arXiv: 2005.11699 (2020).

30. Ivanov, A., and Andrianov, S. "Matrix Lie Maps and Neural Networks for Solving Differential Equations." arXiv preprint arXiv:1908.06088 (2019).

31. Andrianov, S. N. "A role of symbolic computations in beam physics." In International Workshop on Computer Algebra in Scientific Computing (2010, September) pp. 19-30. Springer, Berlin, Heidelberg.

32. Nocedal, J., and Wright, S. Numerical optimization. (2006) Springer Science & Business Media.

33. Tikhonov, A. N., Goncharsky, A. V., Stepanov, V. V., and Yagola, A. G. Numerical methods for the solution of ill-posed problems (Vol. 328). (2013) Springer Science & Business Media.

34. Schwetlick, H., and Schütze, T. "Least squares approximation by splines with free knots. BIT Numerical mathematics." 35(3) (1995), 361-384; Golub G. H. and Pereyra V., "The differentiation of pseudoinverses and nonlinear least squares problems whose variables separate." SIAM J Numer Anal (973), pp 4 3-43

35. Chen, T. C., Han, D. J., Au, F. T., and Tham, L. G. "Acceleration of Levenberg-Marquardt training of neural networks with variable decay rate." In Proceedings of the International Joint Conference on Neural Networks, 2003. (Vol. 3, pp. 1873-1878). IEEE.

36. "Van der Pol equation", Encyclopedia of Mathematics, EMS Press, 2001
37. Lorenz, E. N. "Deterministic nonperiodic flow." Journal of the atmospheric sciences, 20(2) (1963), 130-141.

FIGURES

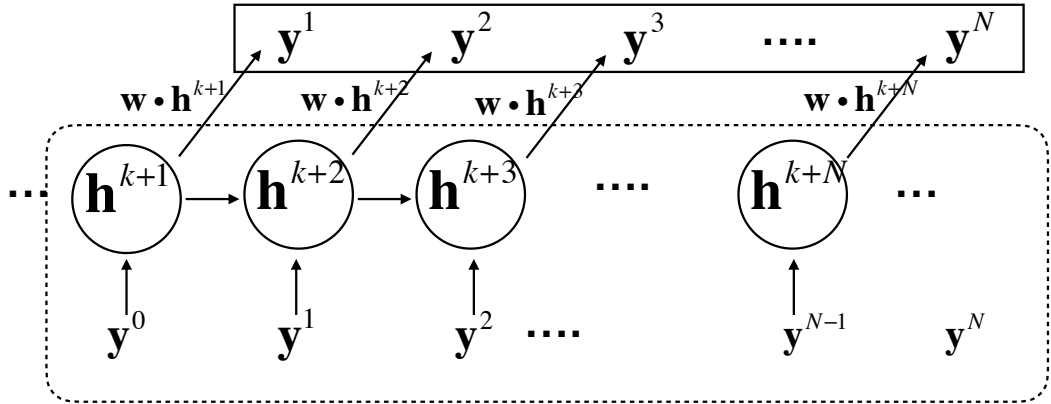

**Fig. 1** An illustration of the typical reservoir readout to be determined by a simple linear regression: as the target signal $\mathbf{y}^n$ is assigned to the input sequence, the hidden states $\mathbf{h}^{n+1}$ are prepared as a fixed sequence which is noted as the things in the round box of dotted line. Then, the sequence of $\mathbf{h}^{n+1}$ is to be transformed into the readout $\mathbf{w} \cdot \mathbf{h}^{n+1}$ only depending on the weight matrix $\mathbf{w}$. Thus, the training process is a simple linear regression to fit the readout to the upcoming target signal $\mathbf{y}^{n+1}$ in the solid box.

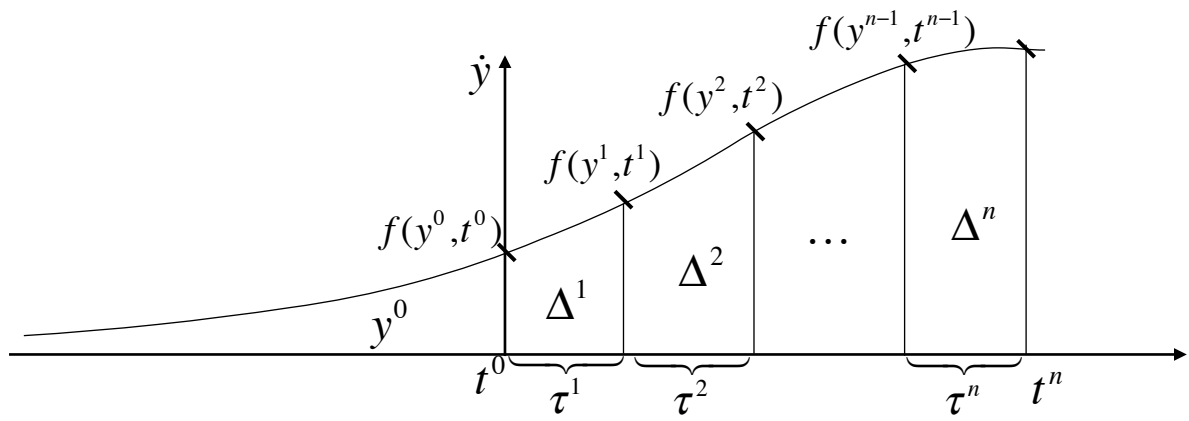

**Fig. 2** The solution of an ODE $d\mathbf{y}/dt = \mathbf{f}(\mathbf{y},t)$ or $\mathbf{f}(\mathbf{y})$ represented as the integral of $\mathbf{f}$ along $t$: each piecewise term $\Delta^n$ by the integration should be the solution with respect to the corresponding initial condition given by another integral up to the initial point, and any contiguous summation of them also satisfies the same logic. Thus, one can think this idea on the concept of invariance led by the basic principle of differential equations.

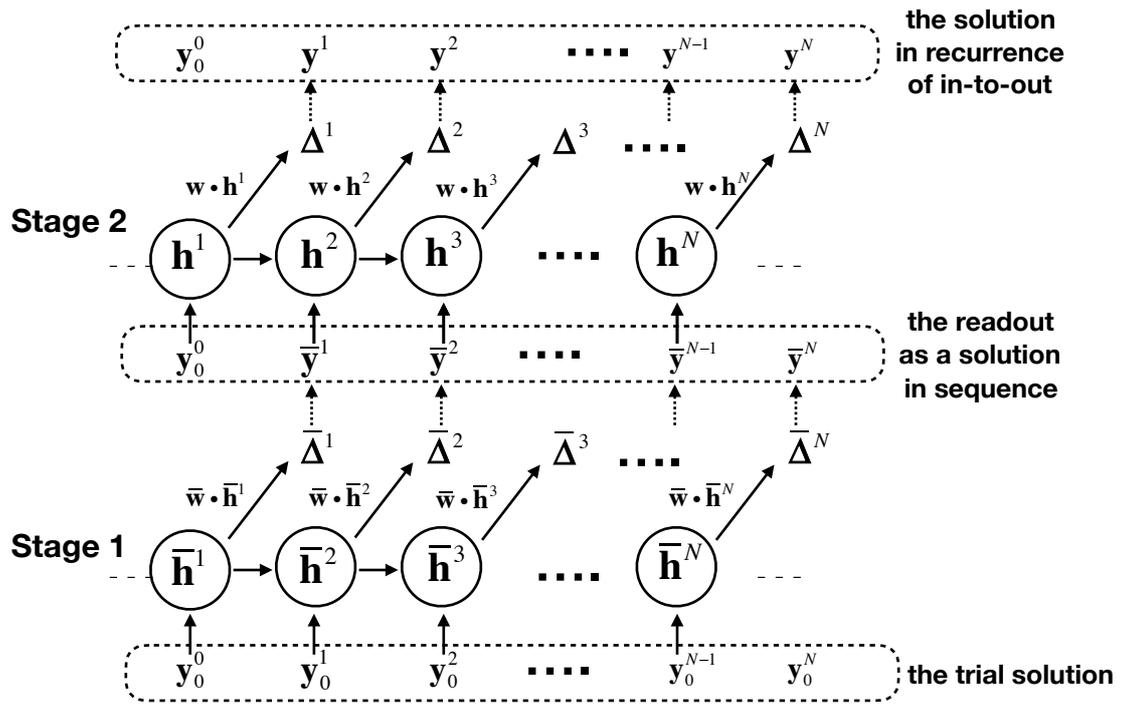

**Fig. 3** The idea of two-pass regression scheme: please care about the notations which the same to Eq. 9 and Eq. 10. One has to be noted that the first stage updates the readout as $\bar{\mathbf{y}}^{n+1} = \bar{\mathbf{y}}^n + \tau^n \bar{\mathbf{w}} \cdot \bar{\mathbf{h}}^{n+1}$, i.e., based on the previous readout, but the second stage applies the input data instead to be $\mathbf{y}^{n+1} = \bar{\mathbf{y}}^n + \tau^n \mathbf{w} \cdot \mathbf{h}^{n+1}$.

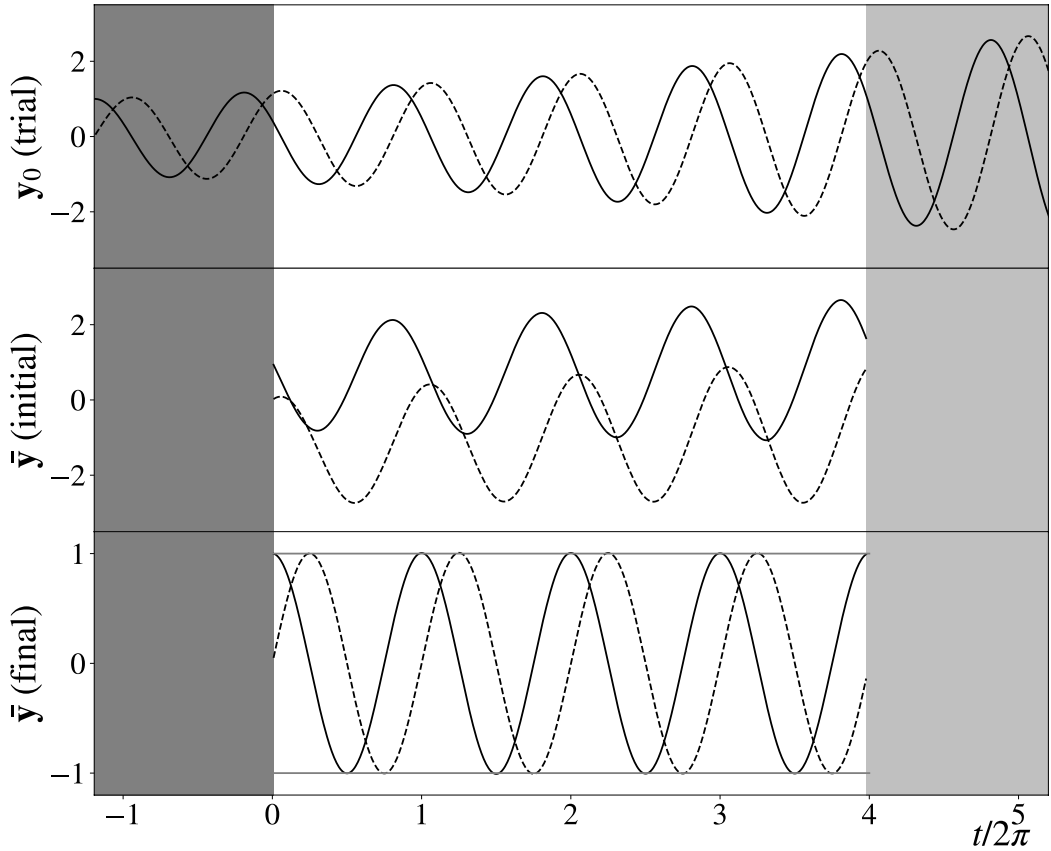

**Fig. 4** Training by the harmonic oscillator equation in the first stage: the trial solution is prepared by the explicit Euler method (upper). Then, the initial state for the regression process is presented after the initial guess of the weight matrix (middle). The shade areas are just dropped for the span of neural solution to be 500 points, and the leading part of 150 points in the darker shade are rejected to "wash out" the transient of the hidden layer [3]. The finial state of regression is represented as accurate sinusoidal waves (lower)

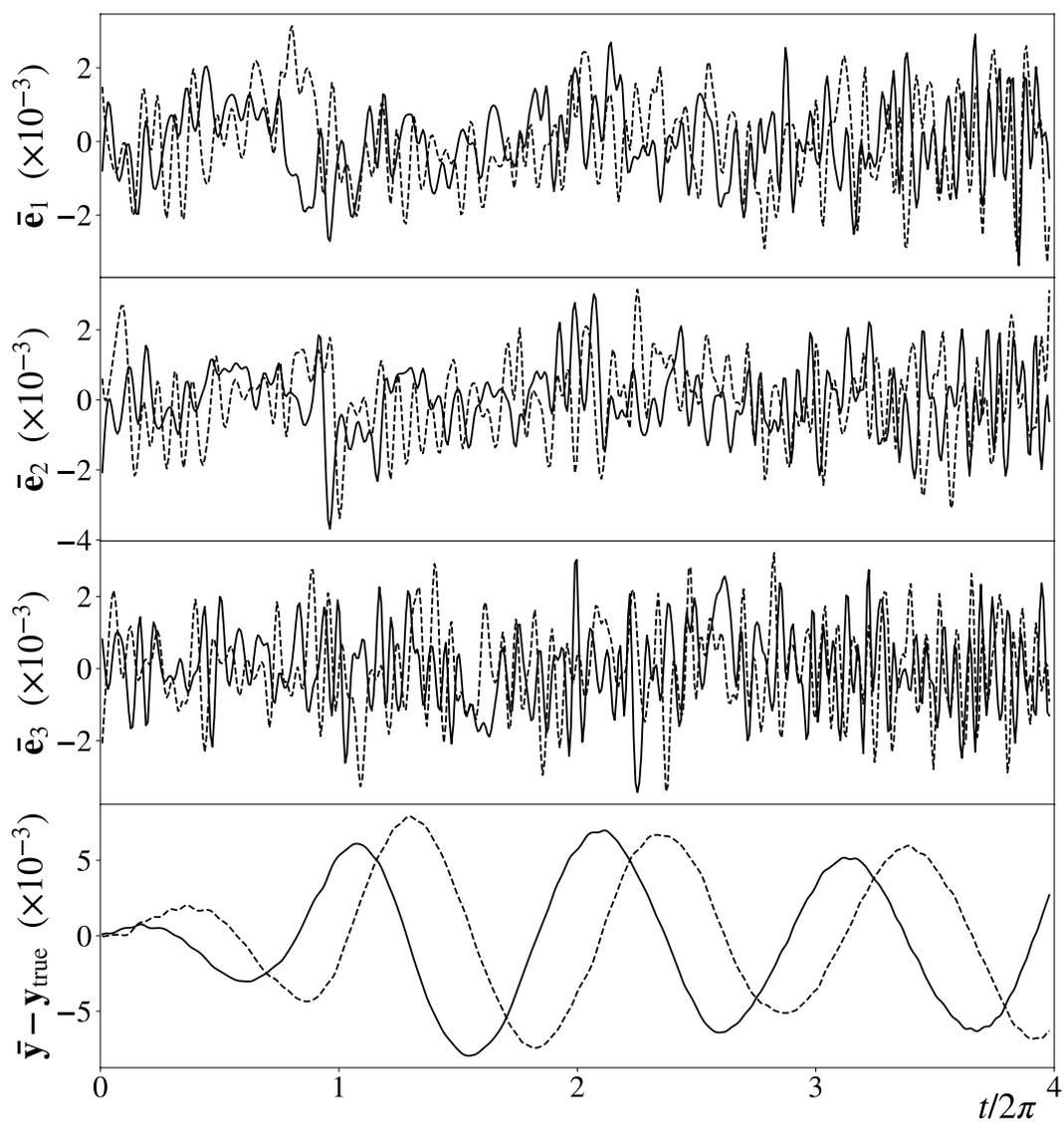

**Fig. 5** The result of regression for the harmonic oscillator in the first stage: the final error vectors per component (upper three) are plotted, as well as the difference from the true solution as small as ~5×10⁻³ (the lowest), where the amplitude is 1.0.

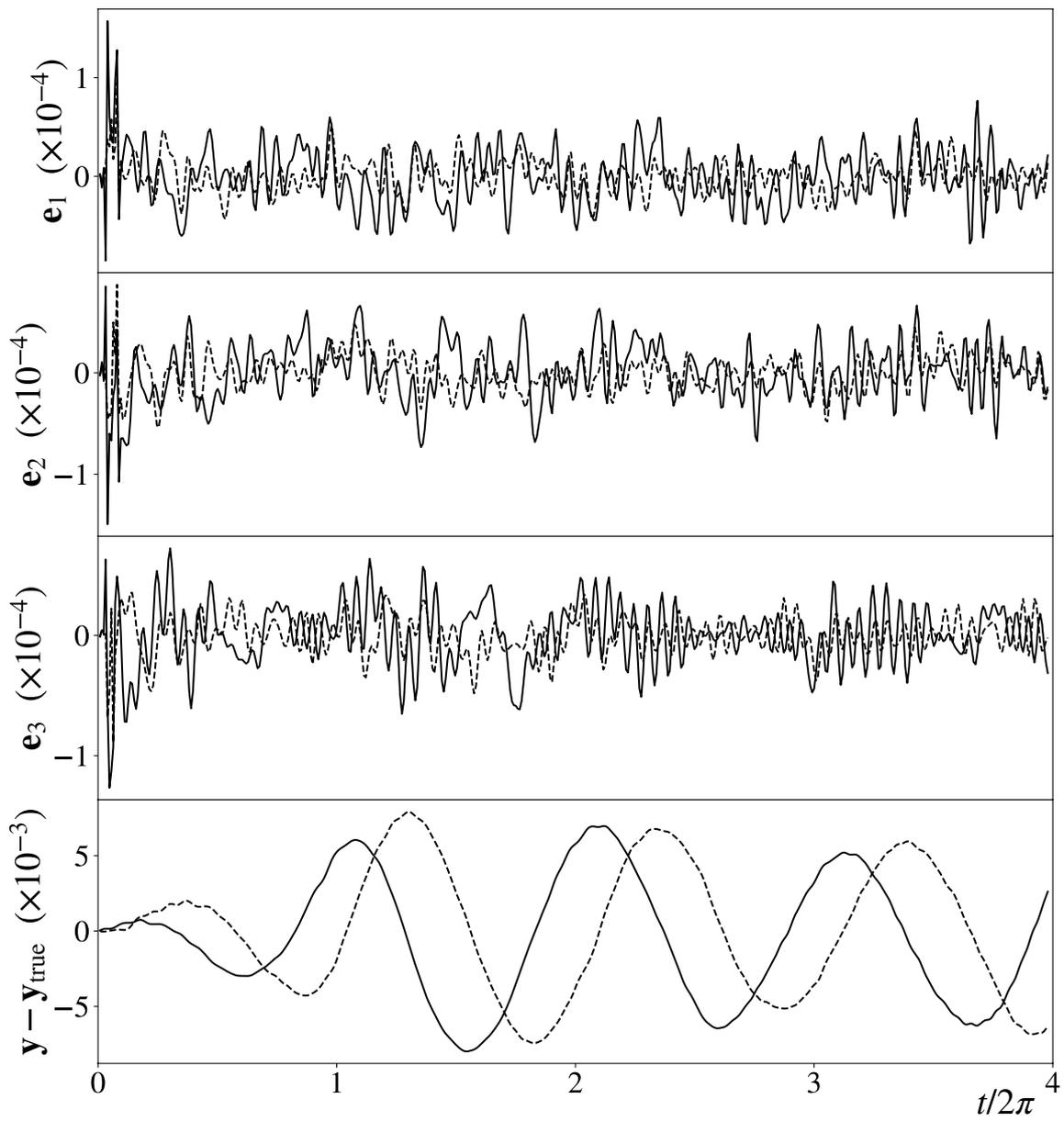

**Fig. 6** The result of regression for the harmonic oscillator in the second stage : almost the same to the first stage

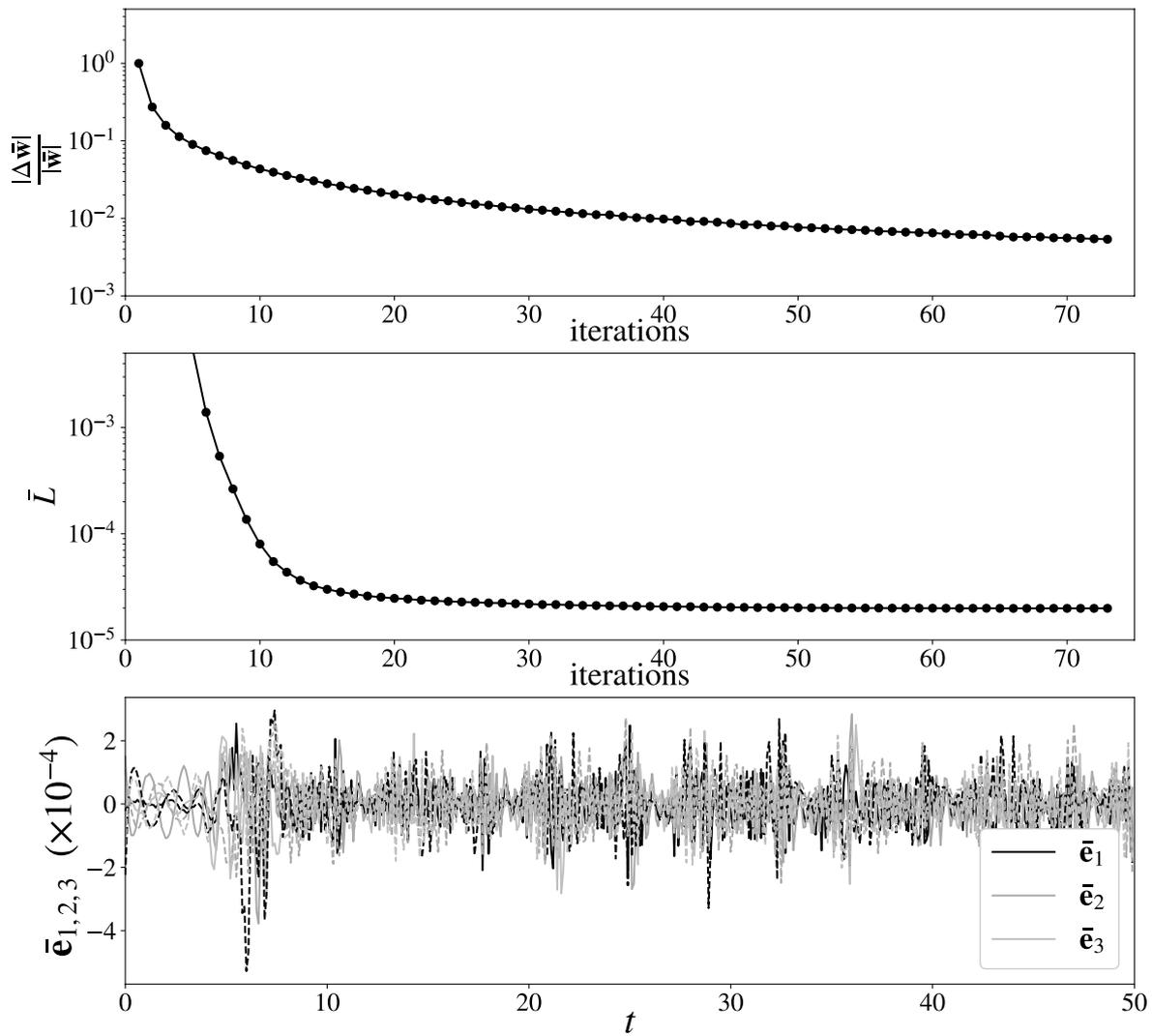

**Fig. 7** The first stage regression process for the van der Pol oscillator : the relative change of the weight matrix versus the number of iterations (upper), the amount of error versus the number of iterations (middle), and the final error vectors (lower). 300 hidden neurons are configured, where their connectivity is 0.1, and the 2-norm of the connection weight is 10.0. The regularization parameter $\lambda$ is fixed as $1 \times 10^{-7}$.

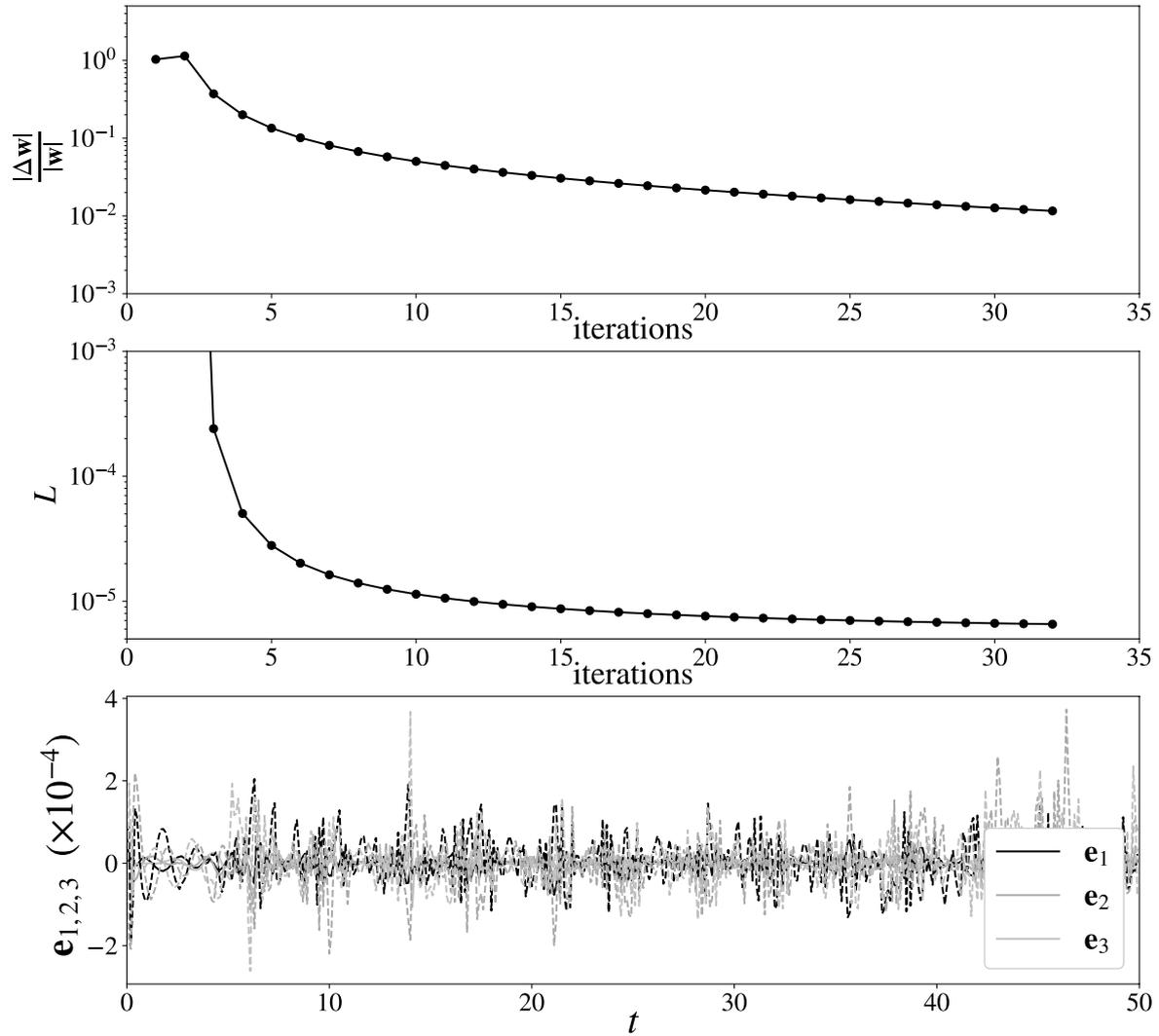

**Fig. 8** The second stage regression process for the van der Pol oscillator : the relative change of the weight matrix versus the number of iterations (upper), the amount of error versus the number of iterations (middle), and the final error vectors (lower)

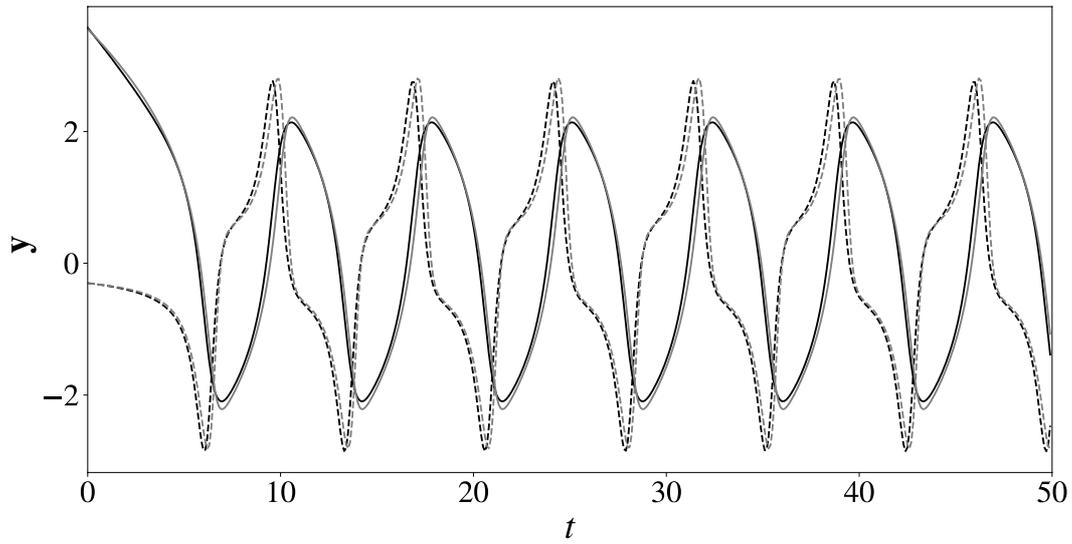

**Fig. 9** The result of the van der Pol Oscillator: $y_1$ is plotted in solid line, and $y_2$ is in dotted line; the trial solution is plotted in grey color just for comparison.

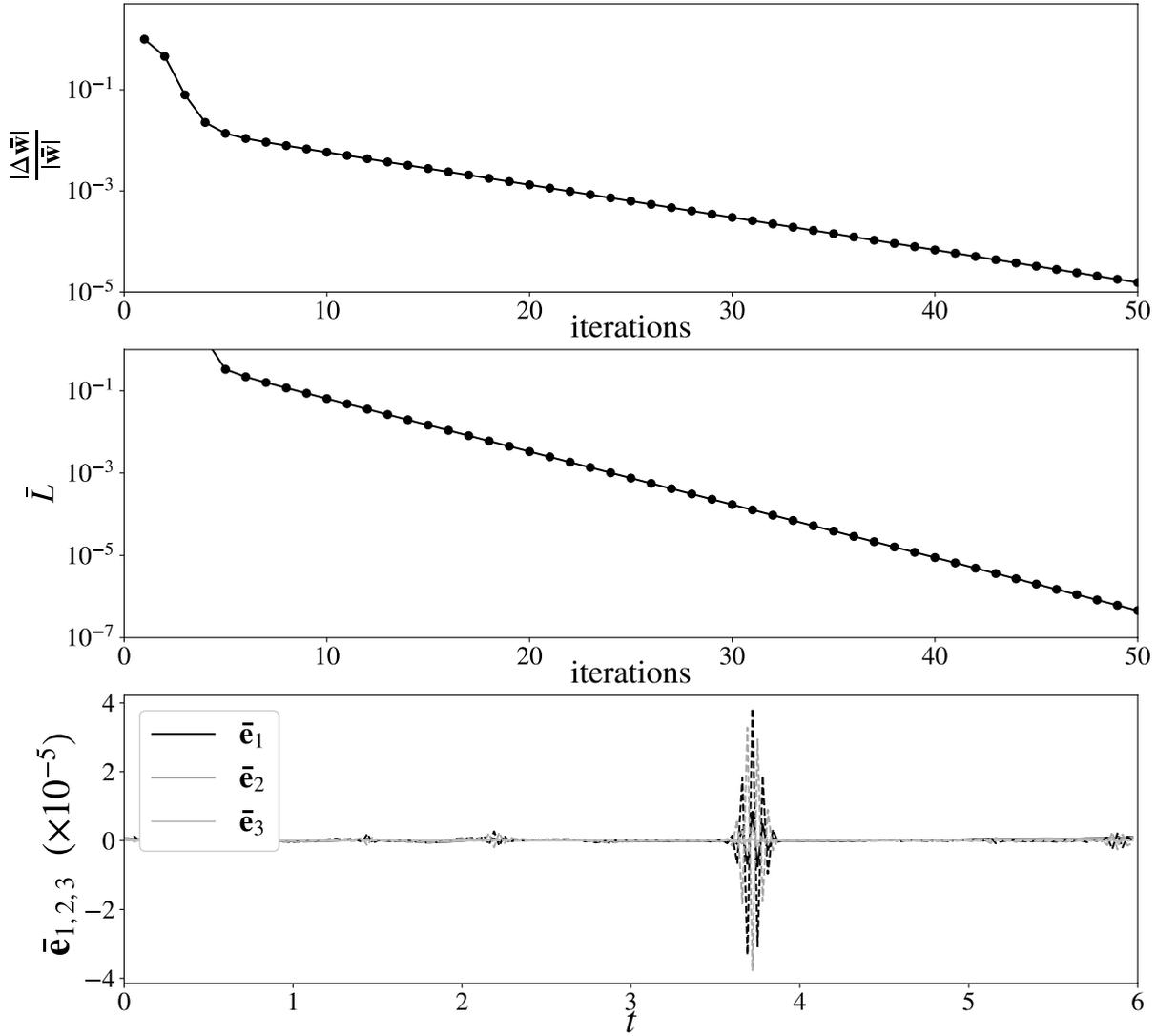

**Fig. 10** The first stage regression process for the Lorenz system : the relative change of the weight matrix versus the number of iteration (upper), the amount of error versus the number of iteration (middle), and the final error vectors (lower). 1500 hidden neurons are configured, where their random connectivity is 0.1, and the 2-norm of the connection weight is 10.0. The regularization parameter $\lambda$ is fixed as $1\times10^{-6}$.

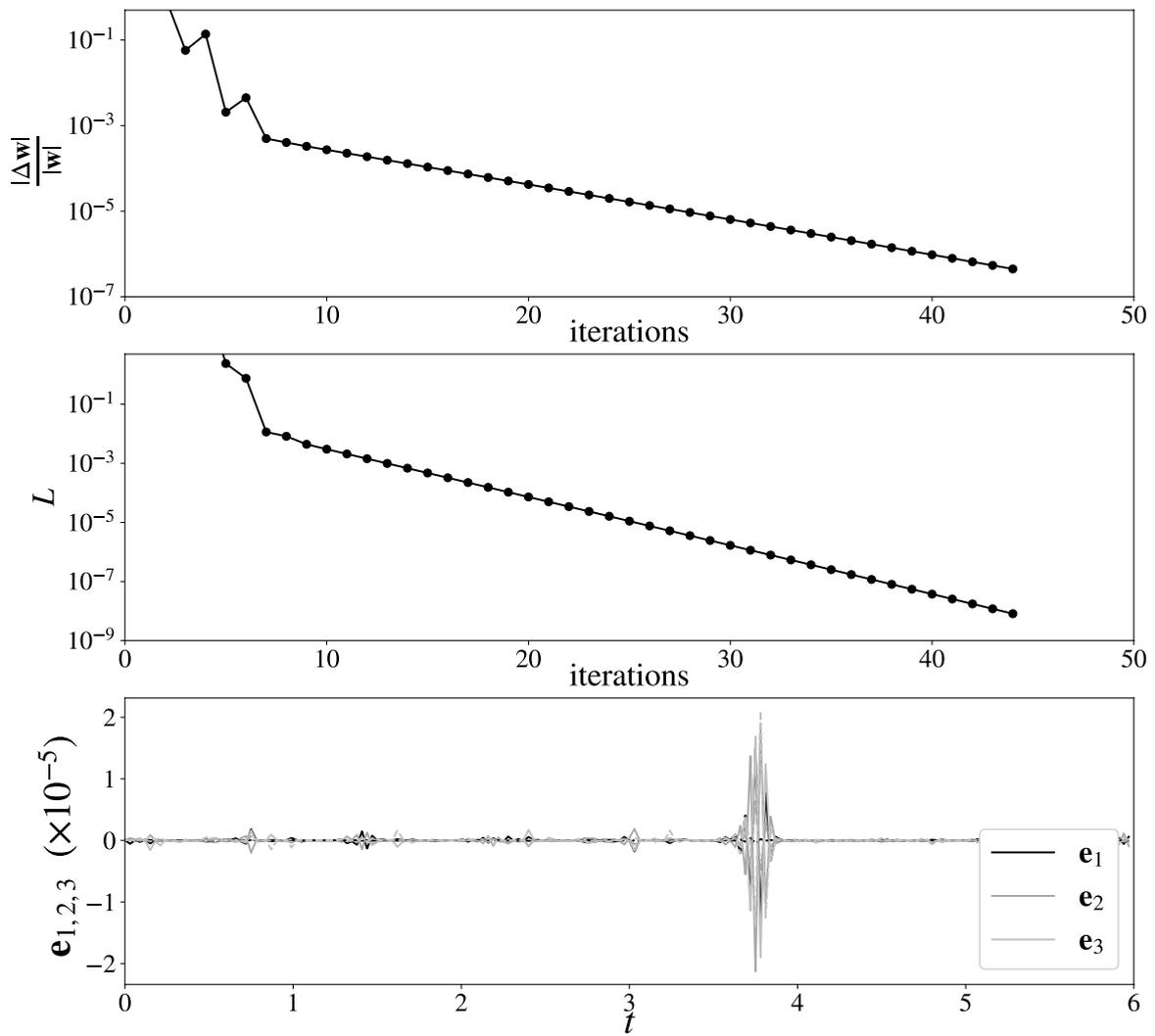

**Fig. 11** The second stage regression process for the Lorenz system: the relative change of the weight matrix versus the number of iterations (upper), the amount of versus the number of iterations (middle), and the final error vectors (lower)

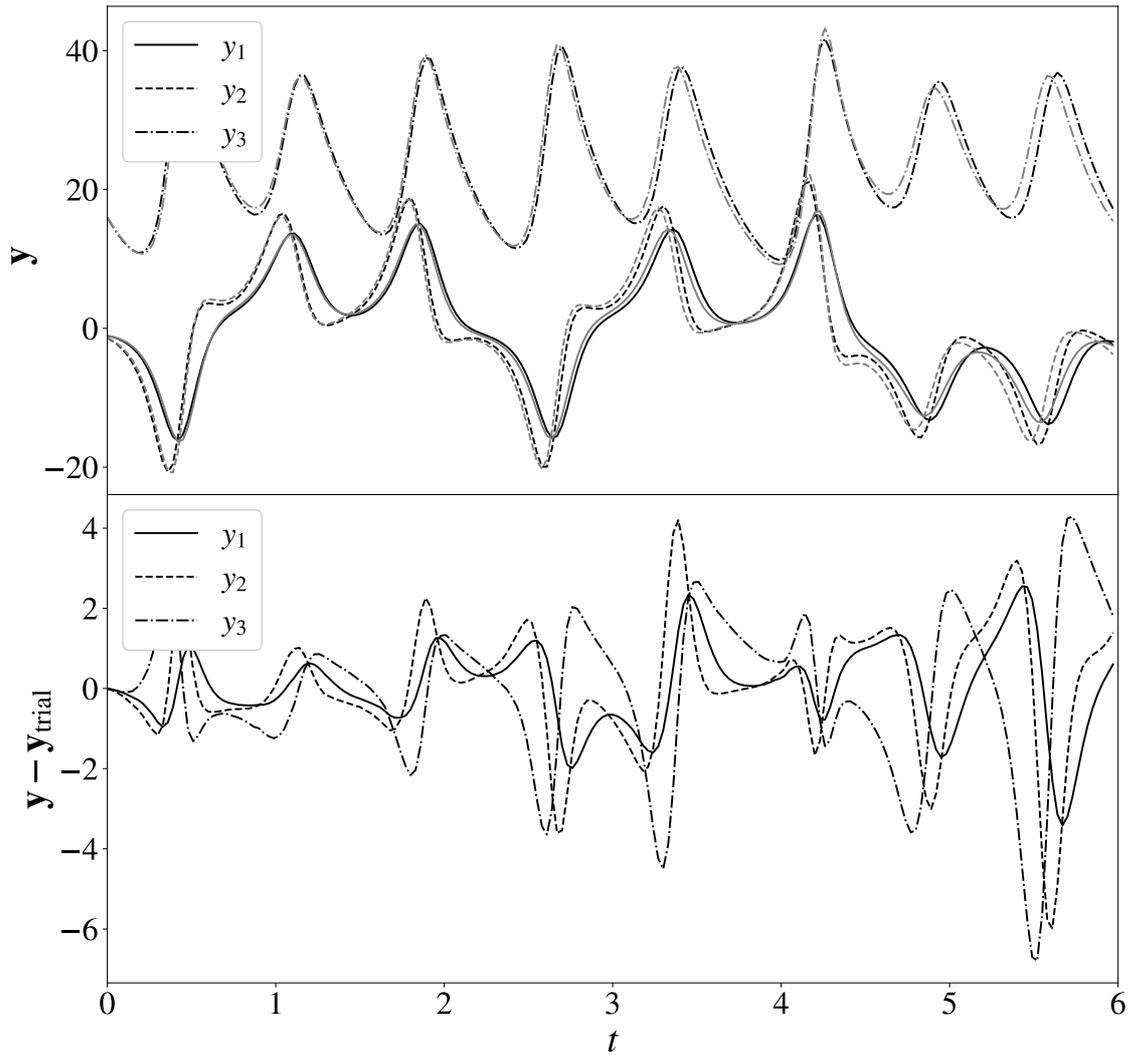

**Fig. 12** The result of the Lorenz system: the final neural output is plotted in solid line as well as the trial solution in grey color just for comparison (upper), and the difference between the result and the trial solution is presented (lower).